\documentclass[10pt,final,journal]{IEEEtran}

\usepackage{amsmath}
\usepackage{amsfonts}
\usepackage{amssymb}
\usepackage{amsthm}
\usepackage{color}
\usepackage{array}
\usepackage{cite}
\usepackage{enumerate}
\usepackage{graphicx}
\usepackage{epstopdf}
\usepackage{epsfig}
\usepackage{footmisc}
\usepackage{amsbsy}
\usepackage{bm}
\usepackage{cases}
\usepackage{multirow}
\usepackage{algorithmic}
\usepackage{comment}
\usepackage{times}
\usepackage{paralist}
\usepackage[ruled]{algorithm2e}
\usepackage{setspace}
\usepackage[breakable]{tcolorbox}
\usepackage{kotex}
\usepackage{caption}
\usepackage{subcaption}

\newcommand{\bmx}{{\bm x}}

\newcommand{\bmh}{{\bm h}}

\newcommand{\set}[1]{\ensuremath{\mathcal #1}}

\begin{document}
\bstctlcite{IEEEexample:BSTcontrol}

\title{Spiking Neural Networks -- \\ Part III: Neuromorphic Communications}

%
%
\author{Author~1, Author~2 and Author~3
\thanks{The authors are with Centre, Department, University, Country (e-mail: firstname.lastname\}@domain).
This work has received funding from Funding Body (Grant)}
}

\author{Nicolas~Skatchkovsky, Hyeryung~Jang and Osvaldo~Simeone
\thanks{The authors are with the Centre for Telecommunications Research, Department of Engineering, King’s College London, United Kingdom. (e-mail: \{nicolas.skatchkovsky, hyeryung.jang, osvaldo.simeone\}@kcl.ac.uk). 
This work has received funding from the European Research Council (ERC) under the European Union's Horizon 2020 Research and Innovation Programme (Grant Agreement No. 725731).}
}


\maketitle

\begin{abstract}
Synergies between wireless communications and artificial intelligence are increasingly motivating research at the intersection of the two fields. On the one hand, the presence of more and more wirelessly connected devices, each with its own data, is driving efforts to export advances in machine learning (ML) from high performance computing facilities, where information is stored and processed in a single location, to distributed, privacy-minded, processing at the end user. On the other hand, ML can address algorithm and model deficits in the optimization of communication protocols. However, implementing ML models for learning and inference on battery-powered devices that are connected via bandwidth-constrained channels remains challenging. This paper explores two ways in which Spiking Neural Networks (SNNs) can help address these open problems. First, we discuss federated learning for the distributed training of SNNs, and then describe the integration of neuromorphic sensing, SNNs, and impulse radio technologies for low-power remote inference. 
\end{abstract}

\section{Introduction}

Machine Learning (ML) and wireless communications are increasingly seen as mutually beneficial, with each technology motivating advances and new applications for the other. On the one end, current advances in ML have been largely fueled by a wider availability of data and by an increase in computing power available on large centralized mixed CPU-GPU clusters. This learning approach, with its reliance on massive amounts of data and computing power, raises accessibility, sustainability, and privacy concerns \cite{strubell2019energyml}. These concerns may be addressed by migrating some of the ML applications to a learning paradigm empowered by wireless among mobile devices. This emerging framework, termed Federated Learning (FL) \cite{wang2019adaptivefl, mcmahan2017fl}, enables collaborative inter-device training without direct exchange of data. 
\begin{figure}
  \centering
  \includegraphics[width=0.9\columnwidth]{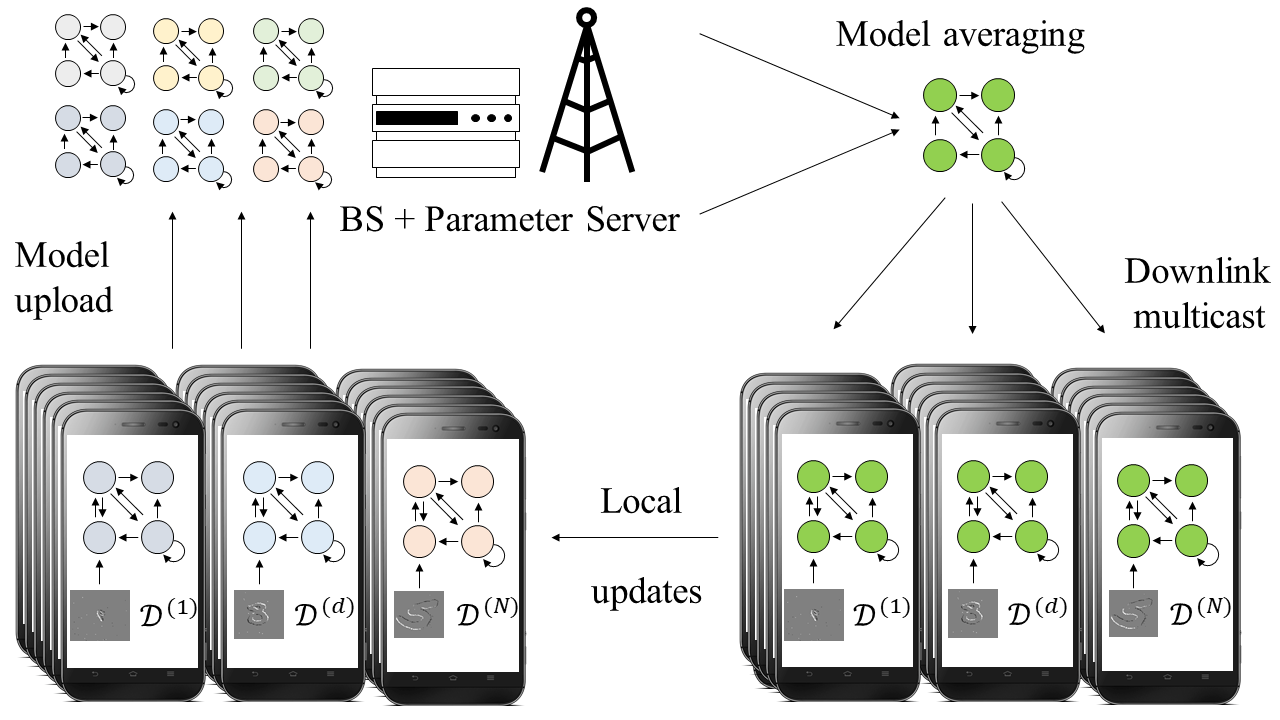}
  \caption{Federated Learning (FL) to collaboratively train on-device SNNs.}
  \label{fig:fl-snn}
  \vspace{-0.5cm}
\end{figure}

On the other hand, novel wireless communication applications, including the deployment of Internet of Things (IoT) networks, pose new problems that challenge the conventional design paradigm that keeps the optimization of sensing, computing, and communication separate. To use spectral resources more effectively, a recent research \cite{oshea2017intro, simeone2018brief, choi2019jssc, raj2019jssc, bourtsoulatze2019jssc} focuses on applying ML to train a communication system in an end-to-end fashion by targeting directly application-level performance criteria. Notable examples include \cite{bourtsoulatze2019jssc}, in which compression and transmission are jointly optimized to minimize the distortion in the transmission of images over wireless links.

As discussed also in Part I and II of this three-part paper \cite{snnreviewpt1, snnreviewpt2}, model architectures that implement standard Artificial Neural Networks (ANNs) may not be the best choice to meet the energy limitations in battery-powered devices. In contrast, Spiking Neural Networks (SNNs) have recently emerged as a low-power alternative thanks to their ability to process information in an event-driven, sparse, and online fashion. SNNs also offer strong synergies with neuromorphic sensors that record \textit{spiking} signals, e.g., from neural prosthetics or silicon retinas, and with impulse radio (IR) communication systems that encode information using the timing of radio pulses \cite{cassidy2008ir, shahshahani2015ir, pepr2016sensor}. IR is a technology of choice for low-power communications. It has been specified by the IEEE 802.15.4z standard, and is investigated for beyond-5G terahertz communication systems \cite{yu2015thzir}.

\begin{figure}
  \centering
  \includegraphics[width=0.9\linewidth]{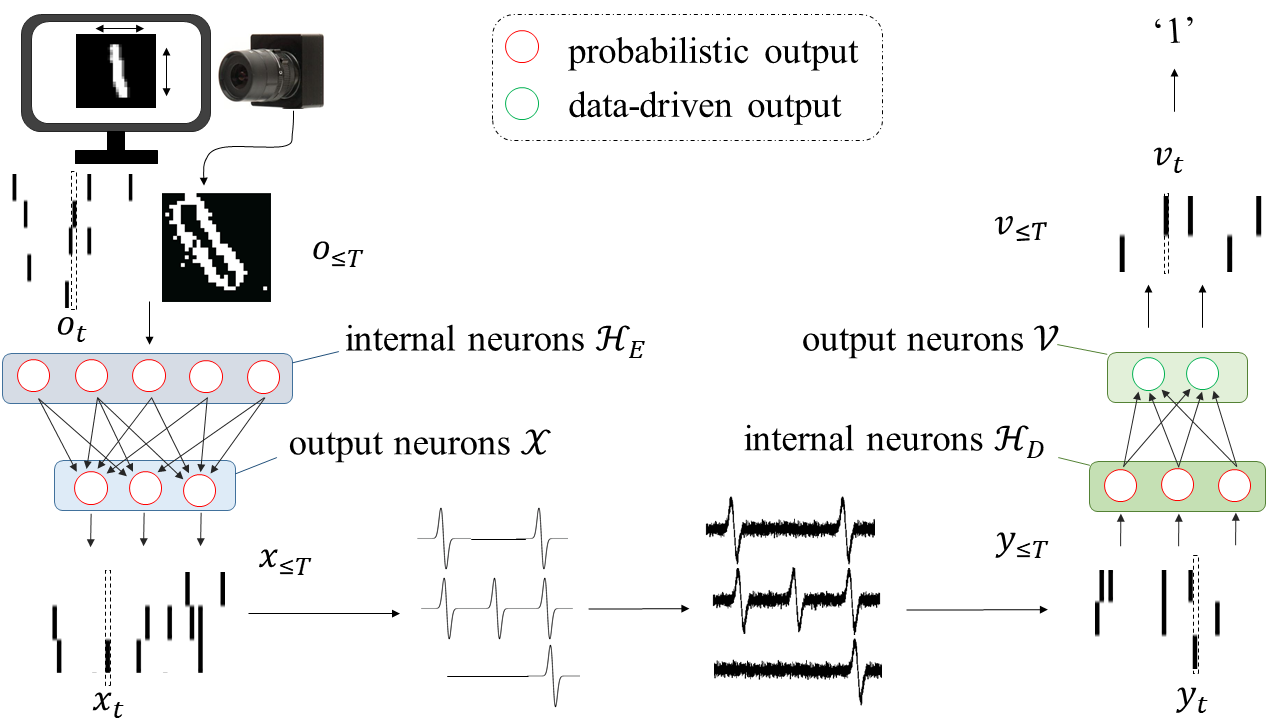}
  \caption{End-to-end neuromorphic sensing, processing, and communications based on spiking sensors, SNNs, and impulse radio (IR).}
  \label{fig:wispike}
\end{figure}

In this paper, we review two applications of SNNs for communication systems. In Sec. \ref{sec:fl-snn}, as illustrated in Fig.~\ref{fig:fl-snn}, we describe an implementation of joint learning via FL that trains on-device SNNs following reference \cite{skatchkovsky2020flsnn}. We discuss some of the novel performance trade-offs that arise when training SNNs instead of ANNs. Then, in Sec. \ref{sec:wispike}, as seen in Fig. \ref{fig:wispike}, we discuss an end-to-end neuromorphic system, for which recording, processing, transmission, and classification are carried out using spiking signals by integrating neuromorphic sensors, SNNs, and IR as in \cite{skatchkovsky2020neurojscc}. We demonstrate significant advantages in terms of time to accuracy and energy efficiency over standard frame-based and digital schemes. 

\section{Communications for Neuromorphic Learning: FL-SNN}
\label{sec:fl-snn}
\subsection{Overview}
\label{sec:fl-snn-overview}

FL has recently emerged as the standard term to describe distributed learning protocols based on the exchange of model, rather than data, information \cite{mcmahan2017fl} with the aim of exploring low-power implementations for on-device training. Unlike standard implementations based on ANNs, we consider the joint training of on-device SNNs via FL. 
The main new challenge in developing FL protocols for the distributed training of SNNs revolves around the online, dynamic, nature of the learning rules for SNNs, which operate over time-encoded inputs as reviewed in \cite{snnreviewpt2}. As a consequence, communication rounds need to be properly scheduled within the finer-grained time scale of the local SNN time steps in order for collaborative training to be effective.

The system, referred to as FL-SNN, consists of $N$ devices communicating through a base station (BS). Each device $d = 1, \dots, D$ holds a different local data set $\mathcal{D}^{(d)}$ that contains a collection of $|\mathcal{D}^{(d)}|$ data points. Following the notation introduced in Part II \cite{snnreviewpt2}, each data point is given by a target spiking signals $\mathbf{x}_{\leq T}^{(d)}$ of duration $T$ samples. The goal of FL is to train a common model based on the global dataset $\mathcal{D} = \cup_{d=1}^{N} \mathcal{D}^{(d)}$ without direct exchange of the data from the local data sets. The training objective of FL-SNN is to solve the problem
\begin{equation}
    \label{eq:general_objective}
     \min_{\theta} F(\mathbf{\theta}) := \frac{1}{\sum_{d=1}^{N}  |\mathcal{D}^{(d)}|}\sum_{d=1}^{D}|\mathcal{D}^{(d)}| F^{(d)}(\mathbf{\theta}),
\end{equation}
where $F(\theta)$ is the global empirical loss over the global data set $\mathcal{D}$; and the local empirical loss at a device $d$ is defined as 
\begin{equation}
\label{eq:local_loss}
F^{(d)}(\mathbf{\theta}) = \frac{1}{|\mathcal{D}^{(d)}|}\sum_{\mathbf{x}_{\leq T}^{(d)} \in \mathcal{D}^{(d)}} f(\mathbf{\theta}, \mathbf{x}_{\leq T}^{(d)}),
\end{equation}
where  $f(\mathbf{\theta}, \mathbf{x}_{\leq T}^{(d)})$ is the loss function measured on target $\mathbf{x}_{\leq T}^{(d)}$ under model parameter vector $\theta$.

Throughout this paper, we adopt the GLM-based SNN model. Referring to  \cite{skatchkovsky2020flsnn} for details, we denote as $\mathcal{X}^{(d)}$ the set of visible neurons in the read-out layer, and as $\mathcal{H}^{(d)}$ the rest of the neurons in the SNN of device $d$, which are also known as hidden neurons. Each neuron $i$ keeps track of membrane potential variable $u_{i,t}^{(d)}$ that evolves as a function of the spikes produced by pre-synaptic neurons through the synaptic weights $\{w_{i,k}^{(d)} \}$, where $k$ is the index of a pre-synaptic neuron. The spiking probability for neuron $i$ at device $d$ conditioned on the previous outputs of all neurons is given as
\begin{align} \label{eq:spike-prob}
    p_{\theta_i^{(d)}}(s_{i,t}^{(d)} = 1| u_{i,t}^{(d)}) = \sigma(u_{i,t}^{(d)}),   
\end{align}
where $\sigma(a) = (1+\exp(-a))^{-1}$ is the sigmoid function and $\theta_i^{(d)}$ is the vector of model parameters for neuron $i$. By \eqref{eq:spike-prob}, the negative log-probability of the spike is given as $-\log p_{\theta_i^{(d)}}(s_{i,t}^{(d)}|u_{i,t}^{(d)}) = \ell( s_{i,t}^{(d)}, \sigma(u_{i,t}^{(d)}) )$, where $\ell(a,b) = -a \log (b) - (1-a) \log (1-b)$ is the binary cross-entropy loss function. 

Accordingly, the loss function $f(\mathbf{\theta}, \mathbf{x}_{\leq T}^{(d)})$ is chosen as the log-loss $f(\mathbf{\theta}, \mathbf{x}_{\leq T}^{(d)}) =  - \log p_{\theta}(\bmx_{\leq T}^{(d)})$, where $p_{\theta}(\bmx_{\leq T}^{(d)}) = \prod_{t=1}^{T} \prod_{i \in \mathcal{X}^{(d)}} p_{\theta_i^{(d)}}(x_{i,t}^{(d)}|u_{i,t}^{(d)}) = \prod_{t=1}^{T} \prod_{i \in \mathcal{X}^{(d)}}\sigma(u_{i,t}^{(d)}) $ is the probability that the read-out layer of visible neurons $\mathcal{X}^{(d)}$ outputs the target signal $\mathbf{x}_{\leq T}^{(d)} \in \mathcal{D}^{(d)}$. The computation of the training loss function requires to average out the stochastic signals $\mathbf{h}_{\leq T}^{(d)}$ produced by the hidden neurons $\mathcal{H}^{(d)}$, which is generally intractable. Using the causally conditioning notation \cite{kramer1998directed} introduced in Part II \cite{snnreviewpt2} and Jensen's inequality, the log-loss is upper bounded as 
\begin{align} \label{eq:local_loss_elbo}
    &f(\mathbf{\theta}, \mathbf{x}_{\leq T}^{(d)}) = - \log \mathbb{E}_{p_{\theta}(\bmh_{\leq T}^{(d)}||\mathbf{x}_{\leq T}^{(d)})} \big[ p_{\theta}(\bmx_{\leq T}^{(d)} || \bmh_{\leq T-1}^{(d)}) \big] \cr 
    & \qquad \quad \leq \mathbb{E}_{p_{\theta}(\bmh_{\leq T}^{(d)}||\mathbf{x}_{\leq T}^{(d)})} \Big[ \sum_{t=1}^T \sum_{i \in \mathcal{X}^{(d)}} \ell\big( x_{i,t}^{(d)}, \sigma(u_{i,t}^{(d)}) \big) \Big].
\end{align}

The key advantage of bound \eqref{eq:local_loss_elbo} is that it decomposes over the time index $t$.
In FL, training is carried iteratively at each device $d$ via gradient descent updates and periodic averaging of the model parameters across devices. The local updates are computed at each local iteration $j = 1, \ldots, J$. In FL-SNN, these iterations are embedded into the finer-grained time scales $t=1, 2,\dots$ at which spikes are processed by the on-device SNNs. Specifically, each local iteration $j$ is applied every $\Delta t \geq 1$ SNN time steps, that is at the end of the interval $[j] = \{t: (j-1)\Delta t + 1 \leq t \leq j \Delta t \}$. The total number of SNN time-steps and the number $J$ of local iterations are hence related as $N \cdot T = J \Delta t$, where $N$ is the number of training examples presented sequentially in an online fashion. Note that Part II of this review presented the local updates for the special case where $\Delta t  = 1$ \cite[Sec.~III.B]{snnreviewpt2}, and we detail below how to generalize the training rule to $\Delta
 t > 1$.    

Every $\Delta J$ local iterations, each device $d$ uploads its current model iterate $\mathbf{\theta}^{(d)}$ to the BS, which computes the centralized averaged parameter 
\begin{equation}
\label{eq:global_update}
    \theta = \frac{1}{\sum_{d=1}^{D} |\mathcal{D}^{(d)}|} \sum_{d=1}^{D} |\mathcal{D}^{(d)}|\theta^{(d)},
\end{equation}
before sending it back to all devices via multicast downlink transmission. Each device $d$ then initializes its local parameter to $\theta$ for the local updates in iteration $j + 1$.

Local updates of $\theta^{(d)}$ are performed at the SNN of each device $d$ by optimizing via gradient descent an upper bound on the local loss \eqref{eq:local_loss_elbo}. The rule follows the online gradient steps described in Part II, applied every $\Delta t$ time steps. Specifically, the $j$-th local update rule for the synaptic weight $w_{i,k}^{(d)}$ between pre-synaptic neuron $k$ and post-synaptic neuron $i$ in the SNN of device $d$ is given as
\begin{align}
\label{eq:snn_update}
    &\Delta w_{i,k}^{(d)}(j) \cr 
    &=
    \begin{cases}
    \sum_{t \in [j]} \big(x_{i,t}^{(d)} - \sigma(u_{i,t}^{(d)}) \big)  \big( \alpha_t \ast s^{(d)}_{k,t} \big), ~~~~~\text{for}~ i \in \mathcal{X}^{(d)}, \cr 
    \sum_{t \in [j]} \bar{e}^{(d)}_t  \big(h_{i,t}^{(d)} - \sigma(u_{i,t}^{(d)}) \big)  \big( \alpha_t \ast s^{(d)}_{k,t} \big), ~\text{for}~ i \in \mathcal{H}^{(d)}, 
    \end{cases}
\end{align}
where we set $s_{i,t}^{(d)} = x_{i,t}^{(d)}$ as the target spiking signal at time $t$ for any visible neuron $i \in \mathcal{X}^{(d)}$; we use the notation $s_{i,t}^{(d)} = h_{i,t}^{(d)}$ for the binary output of any hidden neuron $i \in \mathcal{H}^{(d)}$ at time $t$; $\alpha_t$ is the filter representing the synaptic spike response; $\ast$ is the convolution operator; and the global error signal $\bar{e}^{(d)}_t$ at device $d$ is computed as
\begin{align}
\label{eq:learning_signal}
\bar{e}^{(d)}_t  = \sum_{i \in \set{X}^{(d)}} \ell\big( x_{i,t}^{(d)}, \sigma(u_{i,t}^{(d)})\big).
\end{align}
The resulting algorithm is summarized in \cite{skatchkovsky2020flsnn}.

\subsection{Experiments} 
\label{sec:fl-snn-results}

\begin{figure}
\centering
\includegraphics[width=0.9\columnwidth]{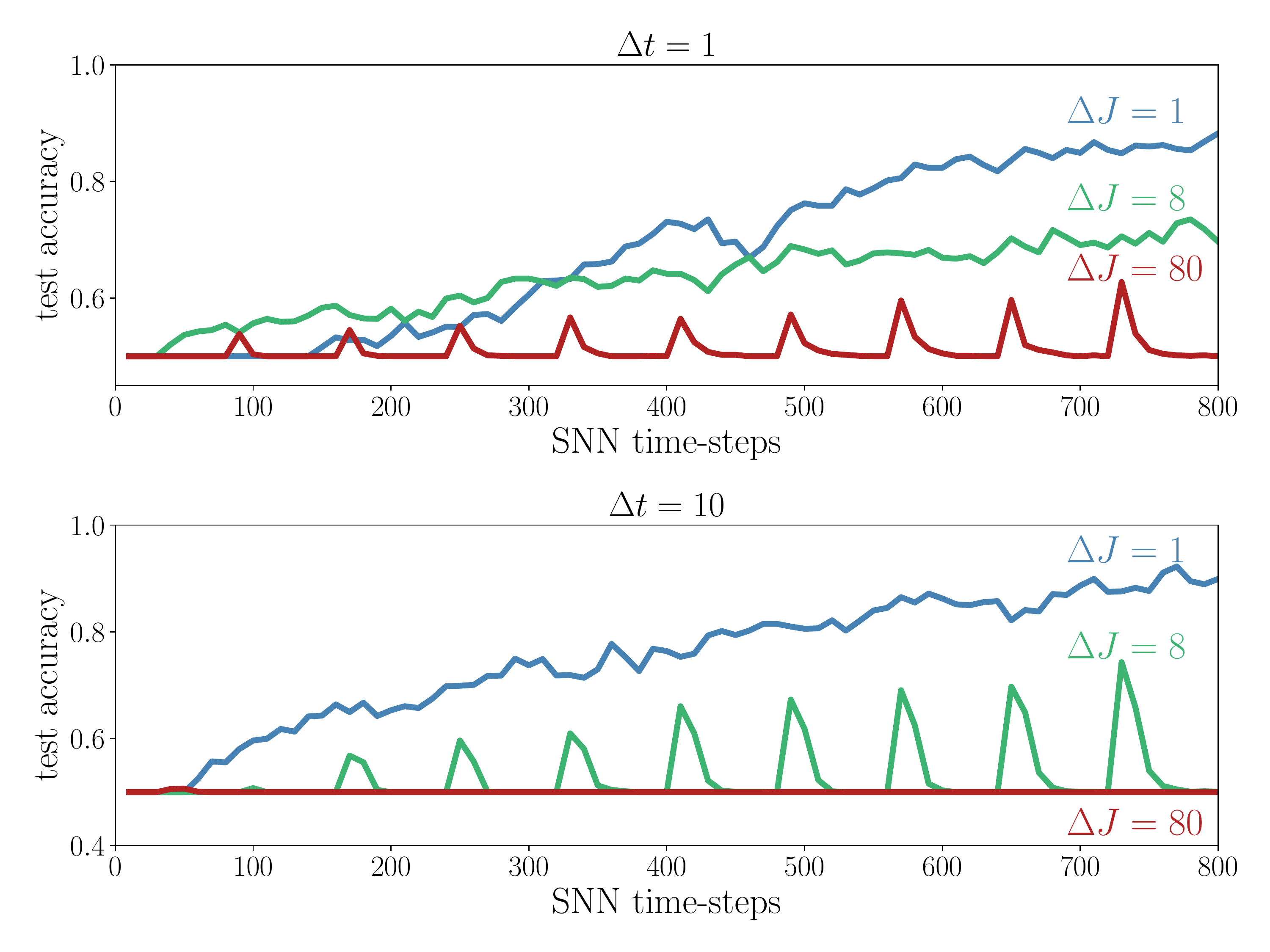}
\vspace{-0.2cm}
\caption{FL with SNNs: Test accuracy with different values of $\Delta J$ during training from a random initialization. Top: $\Delta t = 1$, bottom: $\Delta t = 10$.}
\label{fig:flsnn}
\vspace{-0.4cm}
\end{figure}

We consider the task of classifying handwritten digits displayed to a neuromorphic camera, which were recorded as part of the MNIST-DVS dataset  \cite{serrano2015poker}. We follow the preprocessing procedure detailed in \cite{skatchkovsky2020flsnn}, which yields $676$ input spiking signals with $T = 80$ time-steps from images of size $26 \times 26$.

Digits are split across $D = 2$ devices, with the first device having access only to examples of digit ``1'', and the second having access only to examples from digit ``7''. Neurons in both on-device SNNs are densely connected, and exogenous inputs are connected to all neurons.

In Fig.~\ref{fig:flsnn}, we highlight the impact of the choice of the number $\Delta t$ of time-steps between local updates and of the number $\Delta J$ of local updates between communication rounds. Note that the number of time-steps between communication rounds is $\Delta t \times \Delta J$. The top figure is obtained with $\Delta t = 1$, and the bottom figure with $\Delta t = 10$. We report test accuracy with different values of $\Delta J$ during training from a random initialization, averaged over both devices and three repetitions of the experiment.

With $\Delta t = 1$, local updates have a higher variance, but they are more frequent. More frequent local updates also enable a larger number of communication rounds for a given number of time-steps. With $\Delta t = 10$, updates are averaged over ten time-steps, and are hence less noisy, but occur less often. Fig.~\ref{fig:flsnn} shows that the best performance is obtained with $\Delta t = 10$, which ensures lower-variance updates, and $\Delta J = 1$, which yields frequent communication rounds. Note that, with $\Delta t =1$ and $\Delta J = 1$, training is initially slow since the first communication rounds are impaired by the noisy local updates. 
Furthermore, when $\Delta t = 10$ and $\Delta J = 8$ or $80$, communication is too infrequent to allow for collaborative training, and the test accuracy is limited by overfitting on the local data. As a result, after the improvement caused by a communication round, the performance quickly drops back to yield a classifier with worst-case performance. 
Note that this is due to the fact that testing is done on the test examples selected from both classes. We refer to \cite{skatchkovsky2020flsnn} for a study of the trade-offs in the case of a limited communication budget.

\section{Neuromorphic Learning for Communications: NeuroJSCC}
\label{sec:wispike}
\subsection{Overview}
In this section, we describe a system that processes information, from sensing to inference, in the form of sparse, spiking signals, by integrating neuromorphic sensors, SNNs, and IR. Data is recorded using a neuromorphic low-power sensor, and inference is performed at a remote battery-powered devices to which the sensor is wirelessly connected through a low-power radio interface.

As seen in Fig.~\ref{fig:wispike}, physical signals are recorded using a neuromorphic sensor, and encoded as a vector $\mathbf{o}_{\leq T}$ of $d_o$ spiking signals across $T$ time-steps. For example, the recorded data may be obtained from a DVS camera \cite{Lichtsteiner2006dvs_camera}, to which an object or an activity belonging to a certain class is displayed. The sensed signals are fed as inputs to an encoding SNN that outputs $d_x$ spiking signals $\mathbf{x}_{\leq T}$. We denote as $\mathcal{X}$ the set of (hidden) output neurons, and as $\mathcal{H}^{E}$ the rest of the hidden neurons in the encoding SNN.  We denote the probabilistic mapping of the encoding SNN as $p_{\theta^E}(\mathbf{x}_{\leq T} || \mathbf{o}_{\leq T}) = \prod_{t=1}^{T} p(\mathbf{x}_{\leq t} | \mathbf{x}_{\leq t-1}, \mathbf{o}_{\leq t})$, where superscript $E$ denotes the encoder. The encoded signals $\mathbf{x}_{\leq T}$ are modulated by using $d_x$ parallel IR transmissions, with each spike encoded by an IR waveform such as a Gaussian monopulse. 

The channel output $\mathbf{y}_{t}$ at time $t$ is generally stochastic and depends on its past inputs $\mathbf{x}_{\leq t-1}$ up to time $t - 1$. It includes the effect of waveform generation at the transmitter, intersymbol interference, as well as filtering and thresholding at the receiver. As a result, we assume that the received signal $\mathbf{y}_{t} \in \{0, 1\}^{d_y}$ is binary. The channel can be expressed as the causally conditional distribution  $p(\mathbf{y}_{\leq T} || \mathbf{x}_{\leq T})$.

The received signals $\mathbf{y}_{\leq T}$ are fed to a decoding SNN that produces $d_v$ output spiking signals $\mathbf{v}_{\leq T}$, via a probabilistic mapping $p_{\theta^{D}}(\mathbf{v}_{\leq T}||\mathbf{y}_{\leq T})$ with a learnable parameter vector $\theta^{D}$. The decoding SNN comprises the set $\set{V}$ of visible output neurons in the read-out layer and the set $\mathcal{H}^{D}$ of hidden neurons. The output of the decoding SNN $\mathbf{v}_{\leq T}$ is used for inference, e.g., to predict a class index using standard methods for SNN-based classification \cite{jang19:spm, stanojevic2020FileCB}, such as rate encoding described in Part II.

A data set of input-output pairs $(\mathbf{o}_{\leq T}, \mathbf{v}_{\leq T})$, is available for training. Combining all elements, the joint distribution $p_{\theta}(\mathbf{o}_{\leq T}, \mathbf{v}_{\leq T})$ of the end-to-end system is parameterized by encoder and decoder parameters as $\theta = \{\theta^{E}, \theta^{D} \}$. 
\vspace{-0.2cm}

\subsection{NeuroJSCC}

The task of jointly implementing compression and error-correcting codes is known in the literature as Joint Source-Channel Coding (JSCC). In the system outlined above, JSCC is implemented by using the encoding and decoding SNNs. We hence refer to this solution as \textit{NeuroJSCC}. The end-to-end optimization of the system is carried out by maximizing the log-likelihood that the decoding SNN outputs desired spiking signals $\mathbf{v}_{\leq T}$ in response to a given input $\mathbf{o}_{\leq T}$. Mathematically, in a manner similar to FL-SNN, NeuroJSCC tackles the problem of minimizing the training log-loss \cite{jang19:spm}
\begin{align}
\label{eq:mlproblem}
\min_{\theta} \  - \sum_{(\mathbf{o}_{\leq T},\mathbf{v}_{\leq T}) \in \set{D} }\log p_{\theta}(\mathbf{v}_{\leq T} || \mathbf{o}_{\leq T})
\end{align}
over a training set $\set{D}$.

The log-loss $- \log p_{\theta}(\mathbf{v}_{\leq T} || \mathbf{o}_{\leq T})$ in \eqref{eq:mlproblem} is obtained by averaging over the distribution $p_{\theta}(\mathbf{h}_{\leq T}^E, \mathbf{x}_{\leq T}, \mathbf{y}_{\leq T},  \mathbf{h}_{\leq T}^D  || \mathbf{v}_{\leq T}, \mathbf{o}_{\leq T})$ of signals $\mathbf{h}_{\leq T}^E$ and $\mathbf{h}_{\leq T}^D$ produced by the hidden neurons of encoder and decoder, as well as the channel's inputs and outputs $\mathbf{x}_{\leq T}$ and $ \mathbf{y}_{\leq T}$  as
\begin{align}
    \label{log_likelihood}
    & \log p_{\theta}(\mathbf{v}_{\leq T}|| \mathbf{o}_{\leq T}) = 
        \log \mathbb{E}_{p_{\theta}} \Big[ p_{\theta^{D}}(\mathbf{v}_{\leq T} || \mathbf{y}_{\leq T-1}, \mathbf{h}^{D}_{\leq T-1}) \Big].
\end{align}
To address problem \eqref{eq:mlproblem}, as in the derivation of the learning rule \eqref{eq:snn_update}-\eqref{eq:learning_signal}, we apply stochastic gradient descent on the upper bound obtained via Jensen's inequality (see Part II and \cite{skatchkovsky2020neurojscc} for details)
\begin{align} \label{elbo}
     &- \log  p_{\theta}(\mathbf{v}_{\leq T} || \mathbf{o}_{\leq T}) \cr & \leq \mathbb{E}_{p_{\theta}(\mathbf{h}_{\leq T}^E, \mathbf{x}_{\leq T}, \mathbf{y}_{\leq T}, \mathbf{h}_{\leq T}^D || \mathbf{v}_{\leq T},\mathbf{o}_{\leq T})} \bigg[ \sum_{t=1}^T \sum_{i \in \set{V}} \ell \big( v_{i,t}, \sigma(u_{i,t})\big) \bigg]. \cr
\end{align}

As in \eqref{eq:snn_update}-\eqref{eq:learning_signal}, gradients are approximated via Monte Carlo estimates by drawing samples $\mathbf{h}_{\leq T}^E, \mathbf{x}_{\leq T}, \mathbf{y}_{\leq T}, \mathbf{h}_{\leq T}^D$ of the hidden neurons and channel outputs via the forward operation of the system described by distribution $p_{\theta}(\mathbf{h}_{\leq T}^E, \mathbf{x}_{\leq T}, \mathbf{y}_{\leq T}, \mathbf{h}_{\leq T}^D || \mathbf{v}_{\leq T}, \mathbf{o}_{\leq T})$, yielding the update
\begin{equation}
\label{eq:neurojscc_update}
    \Delta \theta_{i,k} =  \begin{cases}
 \big(v_{i,t} - \sigma(u_{i,t}) \big)  \big( \alpha_t \ast s_{k,t} \big),  &\text{for  $i \in \mathcal{V}$}, \\
\bar{e}_t \big(s_{i,t} - \sigma(u_{i,t}) \big)  \big( \alpha_t \ast s_{k,t} \big), &\text{for $i \in \mathcal{H}^E, \mathcal{X}, \mathcal{H}^D$},
\end{cases}
\end{equation}
where $v_{i,t}$ is the target spiking signal at time $t$ for any visible neuron $i \in \mathcal{V}$; $s_{i,t}$ denotes the binary output of any hidden neuron $i \in \mathcal{H}^E, \mathcal{X}, \mathcal{H}^D$ at time $t$;
and the global error signal is computed as
\begin{align}
\label{eq:neurojscc_ls}
\bar{e}_t  = \sum_{i \in \set{V}} \ell\big( v_{i,t}, \sigma(u_{i,t})\big). 
\end{align}
The resulting algorithm is detailed in \cite{skatchkovsky2020neurojscc}.
\vspace{-0.3cm}

\subsection{Experiments}
\begin{figure}
\centering
\includegraphics[scale=0.27]{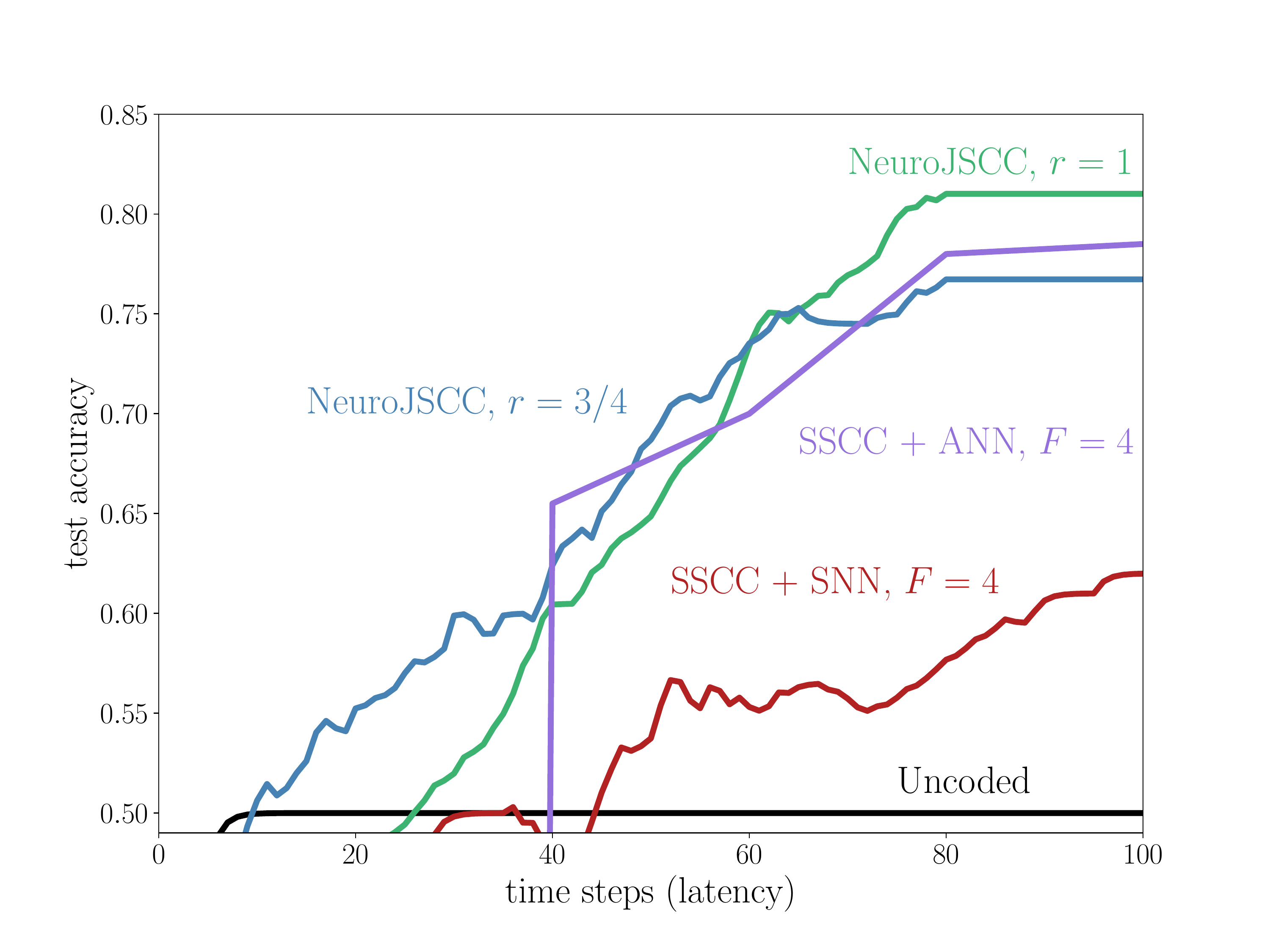}
\vspace{-0.2cm}
\caption{Neuromorphic sensing, communication and inference: Test accuracy as a function of the observation time-steps during inference for Uncoded transmission, Separate Source-Channel Coding and NeuroJSCC schemes (SNR$=-8$ dB).}
\label{fig:acc_per_timestep}
\vspace{-0.6cm}
\end{figure}

In this section, we consider again the problem of classifying between examples of digits ``1'' and ``7'' from the MNIST-DVS dataset. We introduce the communication rate $r = d_x / d_o$ as the ratio between the number of transmitted and input signals at each time-step. We transmit signals through a Gaussian channel, for which we adjust the noise power $\sigma^2$ in order to ensure the same SNR level for all schemes. We define the per-symbol transmission signal-to-noise (SNR) ratio as $\big(|| \mathbf{x}_{\leq T} ||_{1} /(d_{x}T) \big)/\sigma^2$. To obtain binary inputs at the receiver side, the noisy samples are quantized with a hard threshold at $0.5$.
The system of interest is comprised of an encoding SNN with no hidden neurons, and for which all of the $676$ exogeneous inputs are connected to the $d_x = r d_o$ neurons in the output layer. The decoding SNN presents a fully connected architecture with $d_y = d_x$ exogeneous inputs, $N_H^D = d_x$ internal neurons, and $d_v = 2$ output neurons -- one for each class. 

As in \cite{skatchkovsky2020neurojscc}, we compare NeuroJSCC to two schemes. With \textit{Uncoded transmission}, samples are directly transmitted through the channel using On-Off Shift Keying (OOK), yielding a rate $r = 1$. The received are classified using a GLM-based SNN as described in Part II with $N_{H} = 256$ hidden neurons. The second benchmark system is a frame-based \textit{Separate Source-Channel Coding (SSCC)}, which uses state-of-the-art quantization with Vector-Quantization Variational Autoencoder (VQ-VAE) \cite{oord2017discretevae}, and LDPC encoding. The rates of source and channel encoders are chosen to obtain $r = 1$. The scheme divides the input samples into $F$ frames of size $\left \lceil{T/F}\right \rceil$. Frames are classified at the receiver side using either a GLM-based SNN or traditional ANN \cite{deng2019comparison}. We refer to \cite{skatchkovsky2020neurojscc} for further details on the implementation. 

In Fig.~\ref{fig:acc_per_timestep}, we highlight the trade-off between time and accuracy obtained with the different systems for SNR $=-8$ dB. NeuroJSCC is seen to provide a graceful increase in test accuracy with each received sample, while the frame-based scheme is delayed by the necessary time to collect, transmit and process samples in a frame. A smaller transmission results in a faster increase in accuracy, at the cost of a lower performance at convergence. We also observe that uncoded transmission fails to provide better-than-chance classification decisions given the low SNR value.

\begin{figure}
\centering
\includegraphics[scale=0.27]{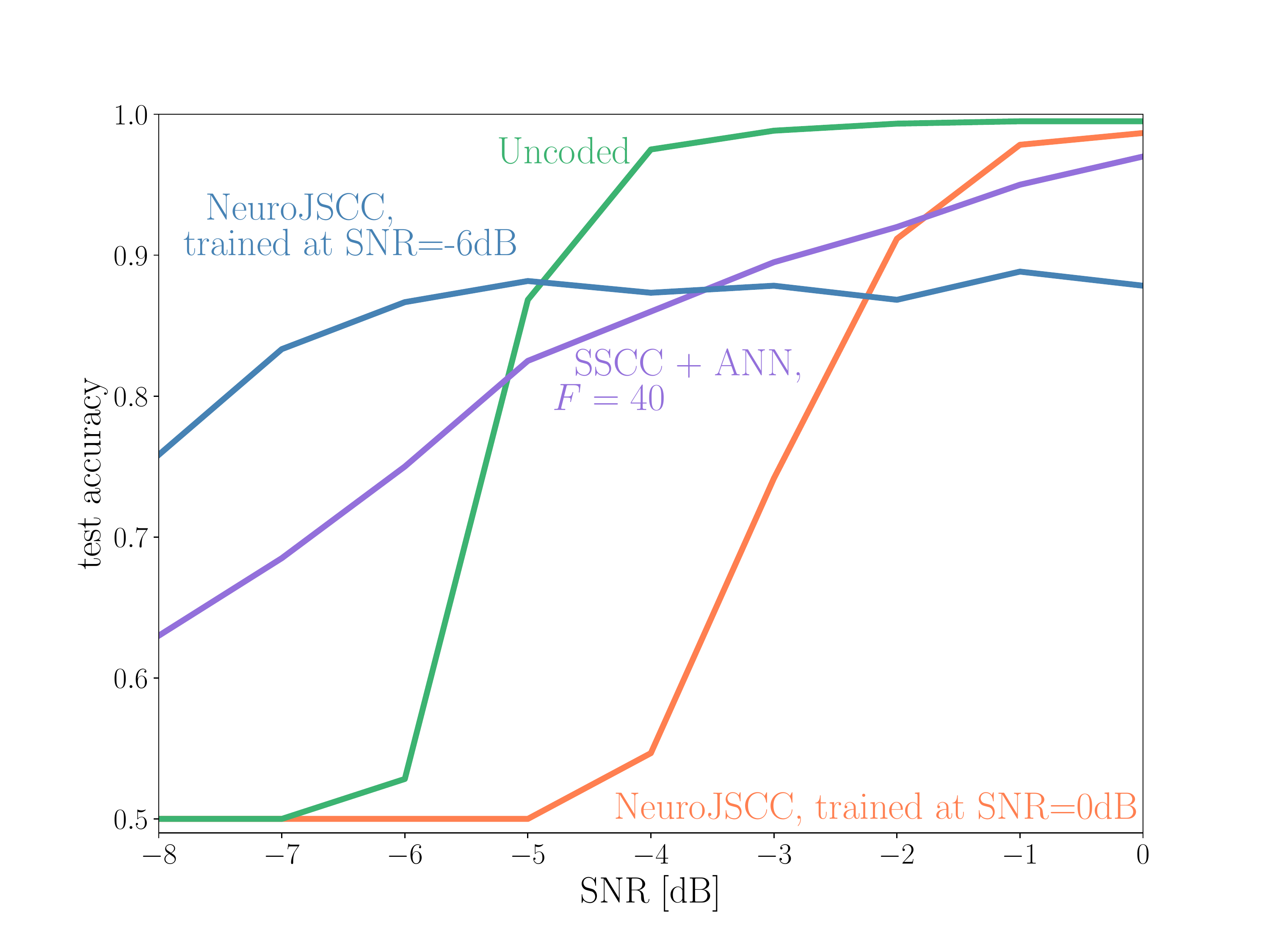}
\vspace{-0.2cm}
\caption{Neuromorphic sensing, communication and inference: Mean test accuracy of NeuroJSCC trained at various SNR levels as a function of the SNR, and comparison with Uncoded and SSCC, obtained by averaging three realizations of each experiment.}
\label{fig:acc_per_snr}
\vspace{-0.6cm}
\end{figure}

In Fig. \ref{fig:acc_per_snr}, we evaluate the test accuracy at convergence obtained for different levels of SNR. NeuroJSCC is trained at a single SNR level, i.e., either $0$ or $-6$ dB. We compare the performance of NeuroJSCC to Uncoded and SSCC, for which training of the auto-encoder is not dependent on the SNR. The accuracy of Uncoded transmission drops sharply at sufficiently low SNR levels. Using Separate SCC with an ANN as classifier proves more robust to low SNR levels. However, NeuroJSCC trained at a SNR of $-6$ dB performs maintains high accuracy at higher SNRs, suggesting the robustness of the scheme to an SNR mismatch.

\vspace{-0.3cm}

\section{Conclusions}
In this paper, the last of a three-part review on SNNs \cite{snnreviewpt1, snnreviewpt2}, we have discussed how SNNs can beneficially be used in synergy with wireless communication systems. On the one hand, SNNs can be integrated with federated learning (FL) to obtain privacy-minded and energy-efficient distributed learning solutions. We have demonstrated by experiments benefits over separate training. On the other hand, neuromorphic sensing and computing can be integrated with impulse radio (IR)-based transmission to ensure low-energy and low-latency remote inference. These solutions exemplify the realm of possibilities offered by the intersection of neuromorphic computing and communications. 
Among the open challenges in neuromorphic FL, we mention here the study of the impact of quantization and, more generally, of communication constraints; the effect of non-stationarity in the statistics of the data, as recently studied in \cite{moraitis2020short} for centralized learning; and an analysis of privacy aspects. As for neuromorphic IR-based communication and learning, we highlight the problem of evaluating the performance on more complex multi-path channels; the study of the impact of non-stationarity in data and channel statistics; and the investigation of multi-access scenarios.
\vspace{-0.2cm}





\bibliographystyle{IEEEtran}
\bibliography{ref}


\end{document}